\documentclass[runningheads]{llncs}
\usepackage{graphicx}
\usepackage{comment}
\usepackage{amsmath,amssymb} 
\usepackage{color}
\usepackage{bm}
\usepackage{mathrsfs}

\usepackage{hyperref}

\def\etal{\emph{et al.}}
\def\ie{\emph{i.e.}}
\def\eg{\emph{e.g.}}

\graphicspath{{Figs/}}

\begin{document}
\pagestyle{headings}
\mainmatter
\def\ECCVSubNumber{7424}  

\title{Self-Paced Deep Regression Forests with Consideration on Underrepresented Examples} 


\titlerunning{Self-Paced DRFs with Consideration on Underrepresented Examples}
%
\author{Lili Pan\inst{1} \and
Shijie Ai\inst{1} \and
Yazhou Ren\inst{1} \and
Zenglin Xu\inst{2,3,1}}
\authorrunning{L. Pan, S. Ai, Y. Ren and Z. Xu}
%
\institute{University of Electronic Science and Technology of China \and
Harbin Institute of Technology, Shenzhen, China \and
Center for Artificial Intelligence, Peng Cheng Lab, Shenzhen, China\\
\email{lilipan@uestc.edu.cn}, asj1995@163.com, yazhou.ren@uestc.edu.cn, zenglin@gmail.com}
\maketitle

\begin{abstract}
Deep discriminative models (\eg~deep regression forests, deep neural decision forests) have achieved remarkable success recently to solve problems such as facial age estimation and head pose estimation.
Most existing methods pursue robust and unbiased solutions either through learning discriminative features, or reweighting samples.
We argue what is more desirable is learning gradually to discriminate like our human beings, and hence we resort to self-paced learning (SPL).
Then, a natural question arises: \emph{can self-paced regime lead deep discriminative models to achieve more robust and less biased solutions?}
To this end, this paper proposes a new deep discriminative model---self-paced deep regression forests with consideration on underrepresented examples (SPUDRFs).
It tackles the fundamental ranking and selecting problem in SPL from a new perspective: fairness.
This paradigm is fundamental and could be easily combined with a variety of deep discriminative models (DDMs).
Extensive experiments on two computer vision tasks, \ie, facial age estimation and head pose estimation, demonstrate the efficacy of SPUDRFs, where state-of-the-art performances are achieved.
\keywords{underrepresented examples, self-paced learning, entropy, deep regression forests.}
\end{abstract}

\section{Introduction}
Deep discriminative models (\eg~deep regression forests, deep neural decision forests) have recently been applied to many computer vision problems with remarkable success.
They compute the input to output mapping for regression or classification by virtue of deep neural networks~\cite{Kontschieder_2015_ICCV,he2016deep,Simonyan2015,shen_deep_2018,chendeepage,chen_using_2017}.
In general, DDMs probably perform better when large amounts of effective training data (less noisy and balanced) is available.
However, such ideal data is hard to collect, especially when large amounts of labels are required.
\begin{figure}[t]
	\centering
	\includegraphics[width=0.85\textwidth]{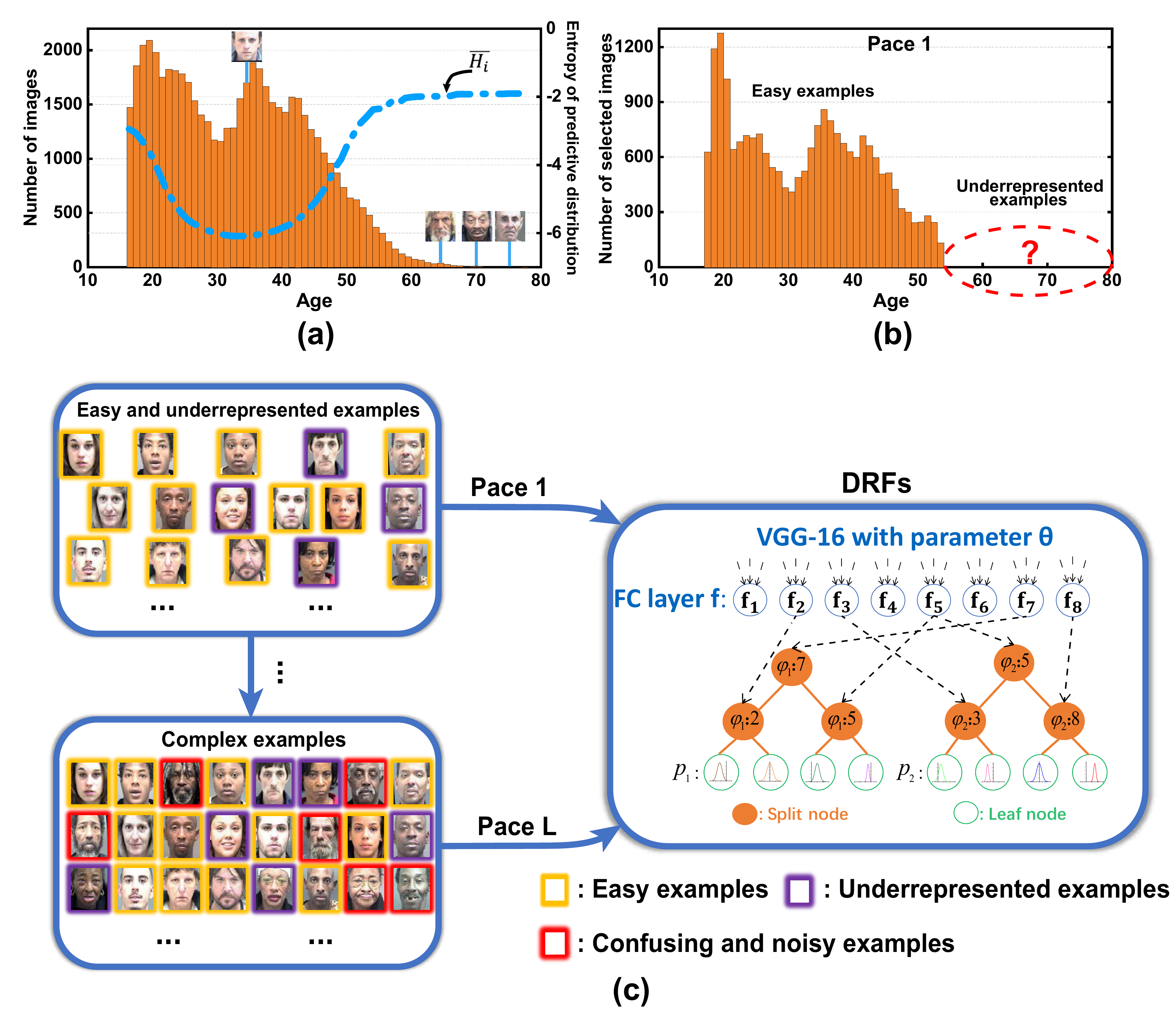}
	\caption{The motivation of considering underrepresented examples in DRFs. \textbf{(a):} The histogram shows the number of face images at different ages, and the average entropy curve represents the predictive uncertainty. We observe the high entropy values correspond to \emph{underrepresented samples}. \textbf{(b):} The histogram of the selected face images at pace 1 in SPL. \textbf{(c):} The proposed new self-paced learning paradigm: easy and underrepresented samples first.}
	\label{Figure1}
\end{figure}

Computer vision literatures are filled with scenarios in which we are required to learn DDMs, not only robust to confusing and noisy examples, but also capable to conquer imbalanced data problem~\cite{zeng2019soft,ren2018learning,cui2019class,khan2019striking,Kortylewski_2019_CVPR_Workshops}.
One typical approach is to learn discriminative features through rather deep neural networks, and feed them into a \emph{cost-sensitive} discriminative function, often with regularization~\cite{Kai2018Deep}.
The other typical approach reweights training samples according to their cost values~\cite{cui2019class,khan2019striking} or gradient directions~\cite{ren2018learning} (\ie~meta learning).
These strategies are unlike our human beings, who lean things gradually---start with easy concepts and build up to complex ones, and can exclude extremely hard ones.
More importantly, we have a sense of \emph{uncertainty} for some samples (\eg~seldom seen) and progressively improve our capability to recognize them.
Thus, the main challenge towards realistic discrimination lies how to mimic our human discrimination system might work.

This line of thinking makes us resort to self-paced learning---a gradual learning regime inspired by the manner of humans~\cite{Kumar2010Self}.
In fact, there are rare studies on the problem of self-paced DDMs.
Then, a natural question arises: \emph{can the self-paced regime lead DDMs to achieve more robust and less biased solutions? }

Motivated by this, we propose a new self-paced learning paradigm for DDMs, which tackles the fundamental ranking and selecting problem in SPL from a new perspective: fairness.
To the best of our knowledge, this is the first work considering \emph{ranking fairness} in SPL.
Specifically, we focus on deep regression forests (DRFs), a typical discriminative method, and propose self-paced deep regression forests with consideration on underrepresented examples (SPUDRFs).
First, by virtue of SPL, our model distinguishes confusing and noisy examples from regular ones, and emphasizes more on ``good'' examples to obtain robust solutions.
Second, our method considers underrepresented examples, which may incur neglect in SPL since visual data is often imbalanced, renderring less bisaed solutions.
Third, we build up a new self-paced learning paradigm: ranking samples on the basis of both likelihood and entropy (predictive uncertainty), as shown in Fig.~\ref{Figure1}, which could be easily combined with a variety of DDMs.

For verification, we apply the SPUDRFs framework on two computer vision problems: (\romannumeral1)  facial age estimation, and (\romannumeral2) head pose estimation.
Extensive experimental results demonstrate the efficacy of our proposed new self-paced paradigm for DDMs.
Moreover, on both aforementioned problems, SPUDRFs almost achieve the state-of-the-art performances.

\section{Related Work}

This section reviews the deep discriminative methods for facial age estimation and head pose estimation, and SPL methods.

\noindent\textbf{Facial Age Estimation.}
DDM based facial age estimation methods, for example~\cite{niu_ordinal_2016,chen_using_2017,gao_age_2018,shen_deep_2018,li2019bridgenet}, employ DNNs to precisely model the mapping from image to age.
Ordinal-based approaches~\cite{niu_ordinal_2016,chen_using_2017} resort to a set of sequential binary queries---each query refers to a comparison with a predefined age, to exploit the inter-relationship (ordinal information) among age labels. 
Improved deep label distribution learning (DLDL-v2)~\cite{gao_age_2018} explores the underlying age distribution patterns to effectively accommodates age ambiguity.
Besides, deep regression forests (DRFs)~\cite{shen_deep_2018} connect random forests to deep neural networks and achieve promising results.
BridgeNet~\cite{li2019bridgenet} uses local regressors to partition the data space and gating networks to provide continuity-aware weights.
The final age estimation result is the mixture of the weighted regression results.
Overall, these DDM based approaches have enhanced age estimation performance largely; however, they plausibly ignore one problem: the interference arising from confusing and noisy examples---facial images with PIE (\ie~pose, illumination and expression) variation, occlusion, misalignment and so forth.

\noindent\textbf{Head Pose Estimation.}
For head pose estimation, Riegler~\cite{riegler2013hough} \etal~utilized convolutional neural networks (CNNs) to learn patch features of facial images and achieved better performance.
In~\cite{huang2018mixture}, Huang \etal~adopted multi-layer perceptron (MLP) networks for head pose estimation and proposed multi-modal deep regression networks to fuse RGB and depth information.
In~\cite{wang2019deep}, Wang \etal~proposed a deep coarse-to-fine network for head pose estimation.
In~\cite{ruiz2018fine}, Ruiz \etal~used a large synthetically expanded head pose dataset to train rather deep multi-loss CNNs for head pose estimation and gained satisfied accuracy.
In~\cite{kuhnke2019deep}, Kuhnke \etal~proposed domain adaptation for head pose estimation, assuming shared and continuous label spaces.
Despite seeing much success, these methods seldom consider the potential problems caused by imbalanced and noisy training data, which may exactly exist in visual problems.

\noindent\textbf{Self-Paced Learning.}
The SPL is a gradual learning paradigm, which builds on the intuition that, rather than considering all training samples simultaneously, the algorithm should be presented with the training data from easy to difficult, which facilitates learning~\cite{Kumar2010Self,meng_theoretical_2017}.
Variants of SPL methods have been proposed recently with varying degrees of success.
For example, in~\cite{jiang2015self}, Zhao \etal~generalized the conventional binary (hard) weighting scheme for SPL to a more effective real valued (soft) weighting manner.
In~\cite{ma2017self}, Ma \etal~proposed self-paced co-training which applies self-paced learning to multi-view or multi-modality problems.
In~\cite{han2017self}, Han \etal~made some efforts on mixture of regressions with SPL strategy, to avoid poorly conditioned linear sub-regressors.
In~\cite{Ren2017RoSR,Ren2020SAMVC}, Ren \etal~introduced soft weighting schemes of SPL to reduce the negative influence of outliers and noisy samples.
In fact, the majority of these mentioned methods can be cast as the combination of SPL and shallow classifiers, where SVM and logistic regressors are usually involved.
In computer vision, due to the remarkable performance of DNNs, some authors have realized SPL may guide DDMs to achieve more robust solutions recently.
In~\cite{ijcai2017}, Li \etal~sought to enhance the learning robustness of CNNs with SPL, and proposed SP-CNNs.
However, \cite{ijcai2017} omits one important problem in the discriminative model: the imbalance of training data.
In contrast to SP-CNNs, our SPUDRFs model has three advantages: (i) it emphasizes ranking fairness (\ie~considering underrepresented examples) in SPL, and hence tends to achieve less biased solutions; (ii) its learning regime is fundamental and can be easily combined with other DDMs, especially the ones with predictive uncertainty; (iii) it creatively explores how SPL can integrate with DMMs with a probabilistic interpretation.

Our work is inspired by the existing works~\cite{jiang2014self,yang2019self} which take the class diversity in the sample selection of SPL into consideration.
Jiang \etal~\cite{jiang2014self} encouraged the class diversity in sample selection at the early paces of self-paced training.
Yang \etal~\cite{yang2019self} defined a metric, named complexity of image category, to measure sample number and recognition difficult jointly, and adopted this measure for sample selection in SPL.
In fact, the aforementioned two methods realize the lack of class diversity in SPL's sample selection may achieve biased solutions since visual data is often imbalanced.
But what causes lack of class diversity is exactly the ranking unfairness as underrepresented examples may often have large loss (particular in DDMs).
Not only that, \cite{yang2019self,jiang2014self} are only suitable for classification, but not regression (with continuous and high dimensional output).
In this paper, we will go further along this direction, aiming to tackle the fundamental problem in SPL: ranking unfairness.

\section{Preliminaries}

In this section, we review the basic concepts of deep regression forests (DRFs)~\cite{shen_deep_2018}.

\noindent \textbf{Deep Regression Tree.} DRFs usually consist of a number of deep regression trees.
A deep regression tree, given input-output pairs $\left\{\mathbf{x}_i, y_i\right\}_{n=1}^N$, where $\mathbf{x}_i\in\mathbb{R}^{D_x}$ and $y_i\in\mathbb{R}$, models the mapping from input to output through DNNs coupled with a regression tree.
A regression tree $\mathcal{T}$ consists of split nodes $\mathcal{N}$ and leaf nodes $\mathcal{L}$~\cite{shen_deep_2018}.
More specifically, each split node $n \in \mathcal{N}$ possesses a split to determine whether input $\mathbf{x}_i$ goes to the left or right subtree; each leaf node $\ell \in \mathcal{L}$ corresponds to a Gaussian distribution $p_{\ell}(y_i)$ with mean $\mu_l$ and variance $\sigma^2_l$.

\noindent \textbf{Split Node.}
Split node has a split function, $s_{n}(\mathbf{x}_i ; \bm{\Theta}) : \mathbf{x}_i \rightarrow[0,1]$, which is parameterized by $\bm{\Theta}$---the parameters of DNNs.
Conventionally, the split function is formulated as $s_{n}(\mathbf{x}_i ; \bm{\Theta})=\sigma\left(\mathbf{f}_{\varphi(n)}(\mathbf{x}_i ; \bm{\Theta})\right)$, where $\sigma(\cdot)$ is the sigmoid function, $\varphi(\cdot)$ is an index function to specify the $\varphi(n)$-th element of $\mathbf{f}(\mathbf{x}_i; \bm{\Theta})$ in correspondence with a split node $n$, and $\mathbf{f}(\mathbf{x}_i; \bm{\Theta})$ denotes the learned deep features.
An example to illustrate the sketch chart of the DRFs is shown in Fig.~\ref{Figure1}, where $\varphi_1$ and $\varphi_2$ are two index functions for two trees.
The probability that $\mathbf{x}_i$ falls into the leaf node $\ell$ is given by:

\begin{equation}
\label{Eq.1}
\omega_\ell( \mathbf{x}_i | \bm{\Theta)}=\prod_{n \in \mathcal{N}} s_{n}(\mathbf{x}_i ; \bm{\Theta})^{[\ell \in \mathcal{L}_{n_l}]}\left(1-s_{n}(\mathbf{x}_i ; \bm{\Theta})\right)^{\left[\ell \in \mathcal{L}_{n_r}\right]},
\end{equation}
where $[\mathcal{H}]$ denotes an indicator function conditioned on the argument $\mathcal{H}$. In addition, $\mathcal{L}_{n_l}$ and $\mathcal{L}_{n_r}$ correspond to the sets of leaf nodes owned by the subtrees $\mathcal{T}_{n_l}$ and $\mathcal{T}_{n_r}$ rooted at the left and right children ${n}_{l}$ and ${n}_{r}$ of node $n$, respectively.

\noindent \textbf{Leaf Node.} For tree $\mathcal{T}$, given $\mathbf{x}_i$,  each leaf node $\ell \in \mathcal{L}$ defines a predictive distribution over $y_i$, denoted by $p_{\ell}(y_i)$.
To be specific, $p_{\ell}(y_i)$ is assumed to be a Gaussian distribution: $\mathcal{N}\left(y_i|\mu_l, \sigma^2_l\right)$.
Thus, considering all leaf nodes, the final distribution of $y_i$ conditioned on $\mathbf{x}_i$ is averaged by the probability of reaching each leaf:
\begin{equation}
\label{Eq.2}
p_{\mathcal{T}}(y_i | \mathbf{x}_i ; \bm{\Theta}, \bm{\pi})=\sum_{\ell \in \mathcal{L}} \omega_\ell( \mathbf{x}_i | \bm{\Theta)} p_{\ell}(y_i),
\end{equation}
where $\bm{\Theta}$ and $\bm{\pi}$ represent the parameters of DNNs and the distribution parameters $\left\{\mu_l,\sigma^2_l\right\}$, respectively.
It can be viewed as a mixture distribution, where $\omega_\ell( \mathbf{x}_i | \bm{\Theta)}$ denotes mixing coefficients and $ p_{\ell}(y_i)$ denotes the Gaussian distributions associated with the $\ell^{th}$ leaf node.
Note that $\bm{\pi}$ varies along with tree $\mathcal{T}_k$, and thus we rewrite it as $\bm{\pi}_k$ below.

\noindent \textbf{Forests of Regression Trees.}
Since a forest comprises a set of deep regression trees $\mathcal{F}=\left\{\mathcal{T}_1,...,\mathcal{T}_k\right\}$, the predictive output distribution, given $\mathbf{x}_i$, is obtained by averaging over all trees:
\begin{equation}
\label{Eq.3}
p_{\mathcal{F}}\left(y_i|\mathbf{x}_i,\bm{\Theta},\bm{\Pi} \right)
=
\frac{1}{K}\sum_{k=1}^K p_{\mathcal{T}_k}\left(y_i|\mathbf{x}_i, \bm{\Theta}, \bm{\pi}_k\right),
\end{equation}
where $K$ is the number of trees and $\bm{\Pi}=\left\{\bm{\pi}_1,...,\bm{\pi}_K\right\}$.
$p_{\mathcal{F}}\left(y_i|\mathbf{x}_i,\bm{\Theta},\bm{\Pi} \right)$ can be viewed as the likelihood that the $i^{th}$ sample has output $y_i$.

\section{Self-Paced DRFs with Consideration on Underrepresented Examples}
The problems in training DDMs for visual tasks arise from: (\romannumeral1) the noisy and confusing examples, and (\romannumeral2) the imbalance of training data.
Intuitively inspired by the gradual learning manner of humans, we resort to self-paced learning and explore whether the DDMs, by virtue of SPL, tend to achieve more robust solutions.
Perhaps not easily, in existing SPL, we observe ranking unfairness, as shown in Fig.~\ref{Figure1}.
Motivated by this observation, we propose SPUDRFs, which starts learning with easy yet underrepresented examples, and build up to complex ones.
Such a paradigm avoids overlooking the ``minority'' of training samples, leading to less biased solutions.

\subsection{Underrepresented Examples}
\label{Uncertainty}
Underrepresented examples mean ``minority'', as which the examples with similar or the same labels are scarce.
Unsurprisingly, we observe that they may incur unfairness treatment in the early paces of SPL (see Fig.~\ref{Figure1}(b)), due to imbalanced data distribution.
The underrepresented level could be measured by predictive uncertainty.
Given the sample $\mathbf{x}_i$, its predictive uncertainty is formulated as the entropy of its predictive output distribution $p_{\mathcal{F}}\left(y_i|\mathbf{x}_i,\bm{\Theta},\bm{\Pi} \right)$:
\begin{equation}
\label{Eq.4}
H\left [p_{\mathcal{F}}\left(y_i|\mathbf{x}_i,\bm{\Theta},\bm{\Pi} \right)\right] = \frac{1}{K}\sum^K_{k=1}H\left [p_{\mathcal{T}_k}\left(y_i|\mathbf{x}_i,\bm{\Theta}, \bm{\pi}_k \right)\right],
\end{equation}
where $H\left[ \cdot\right ]$ denotes entropy, and the entropy corresponds to the $k^{th}$ tree is:
\begin{equation}
\label{Eq.5}
H\left [p_{\mathcal{T}_k}\left(y_i|\mathbf{x}_i,\bm{\Theta}, \bm{\pi}_k \right)\right] = -\int p_{\mathcal{T}_k}\left(y_i|\mathbf{x}_i,\bm{\Theta}, \bm{\pi}_k \right)\ln p_{\mathcal{T}_k}\left(y_i|\mathbf{x}_i,\bm{\Theta}, \bm{\pi}_k \right) dy_i,
\end{equation}
The large the entropy is, the more uncertain the prediction should be, \ie, the more underrepresented the sample is.
Considering underrepresented samples can be interpreted as adequately utilizing the ``information'' inherent in such examples in SPL training.

As previously discussed, $p_{\mathcal{T}_k}\left(y_i|\mathbf{x}_i; \bm{\Theta}, \bm{\pi}_k\right)$ is a mixture distribution, taking the form $\sum_{\ell \in \mathcal{L}} \omega_\ell( \mathbf{x}_i | \bm{\Theta)} p_{\ell}(y_i)$, where $\omega_\ell( \mathbf{x}_i | \bm{\Theta)}$ denotes mixing coefficients and $p_{\ell}(y_i)$ denotes the Gaussian distribution associated with the $\ell
^{th}$ leaf node.
In Eq.~(\ref{Eq.5}), the integral of mixture of Gaussians is non-trivial. Monte Carlo sampling provides a way to calculate it, but incurs large computational cost~\cite{huber2008entropy}.
Here, we use the lower bound of this integral to approximate its true value:
\begin{equation}
\label{Eq.6}
H\left [p_{\mathcal{T}_k}\left(y_i|\mathbf{x}_i,\bm{\Theta}, \bm{\pi}_k \right)\right]\approx\frac{1}{2}\sum_{\ell\in\mathcal{L}}\omega_\ell(\mathbf{x}_i|\mathbf{\Theta})\left[\ln \left(2\pi \sigma_\ell^2\right)+1\right].
\end{equation}
The underrepresented examples are often scarce, and have not been treated fairly, resulting in large prediction uncertainty (\ie~entropy).

\subsection{ Objective Function}

Rather than considering all the samples simultaneously, our proposed SPUDRFs are presented with the training data in a meaningful order, that is, easy and underrepresented examples first.
Specifically, we define a latent variable $v_i$ that indicates whether the $i^{th}$ sample is selected $(v_i = 1)$ or not $(v_i = 0)$ depending on how easy and underrepresented it is for training.
Our objective is to jointly maximize the log likelihood with
respect to DRFs' parameters $\bm{\Theta}$ and $\bm{\Pi}$, and learn the latent selecting variables $\mathbf{v}=\left(v_1,...,v_N\right)^T$.
We prefer to select the underrepresented examples, which probably have higher predictive uncertainty (\ie~entropy), particularly in the early paces.
It builds on the intuition that the underrepresented examples may incur neglect since they are the ``minority'' in training data.
Therefore, we maximize a self-paced term regularized likelihood function, meanwhile considering predictive uncertainty,
\begin{equation}
\label{Eq.7}
\max_{\bm{\Theta},\bm{\Pi}, \mathbf{v}} \sum_{i=1}^{N} v_{i} \left \{ \log p_{\mathcal{F}}\left(y_i|\mathbf{x}_i,\bm{\Theta},\bm{\Pi} \right) + \gamma H_i \right \}  + \lambda\sum_{i=1}^N v_i ,
\end{equation}
where $\lambda$ is a parameter controlling the learning pace, $\lambda>0$, $\gamma$ is the parameter imposing on entropy, and $H_i$ denotes the predictive uncertainty of the $i^{th}$ sample, as previously discussed in Sec.~\ref{Uncertainty}.
When $\gamma$ decays to 0, the objective function is equivalent to the log likelihood function with respect to DRFs' parameters $\bm{\Theta}$ and $\bm{\Pi}$.
Eq.~(\ref{Eq.7}) indicates each sample is weighted by $v_i$, and whether $\log p_{\mathcal{F}}\left(y_i|\mathbf{x}_i,\bm{\Theta},\bm{\Pi} \right) + \gamma H_i>-\lambda$ determines
the $i^{th}$ sample is selected or not.
That is, the sample with high likelihood value or high predictive uncertainty may be selected.
The optimal $v_i^*$ is:
\begin{align}
\label{Eq.8}
v_i^* = \left\{ \begin{array}{ll}
1 & \textrm{if $\log p_{\mathcal{F}i} + \gamma H_i > -\lambda$}\\
0 & \textrm{otherwise}
\end{array} \right.,
\end{align}
where $p_{\mathcal{F}}\left(y_i|\mathbf{x}_i,\bm{\Theta},\bm{\Pi} \right)$ is written as $ p_{\mathcal{F}i}$ for simplicity.

One might argue the noisy and hard examples tend to have high predictive uncertainty also, rendering being selected in the early paces.
In fact, from Eq.~(\ref{Eq.8}), we observe whether one sample is selected is determined by both its predictive uncertainty and the log likelihood of being predicted correctly.
The noisy and hard examples probably have relatively large loss \ie~low log likelihood, avoiding being selected at the very start.

Iteratively increasing $\lambda$ and decreasing $\gamma$, samples are dynamically involved in the training of DRFs, starting with easy and underrepresented examples and ending up with all samples.
Note every time we retrain DRFs, that is, maximizing Eq.~(\ref{Eq.7}), our model is initialized to the result of the last iteration.
As such, our model is initialized progressively by the result of the previous pace---adaptively calibrated by ``good'' examples.
This also means we place more emphasis on easy and underrepresented examples rather than confusing and noisy ones.
Thus, SPUDRFs are prone to have more robust and less biased solutions since we adequately consider the underrepresented examples.

\noindent \textbf{Mixture Weighting.}
In the previous section, we adopt a hard weighting scheme to assign data points to paces, in which one sample is either selected $(v_i=1)$ or not $(v_i=0)$.
Such a weighting scheme appears to be less accurate as it omits the importance of samples.
Hence, we adopt a mixture weighting scheme~\cite{jiang2014easy}, where the selected samples are weighted by its importance, ling in the range $0\leq v_i \leq 1$.
The objective function with mixture weighting is defined as:
\begin{equation}
\label{Eq.9}
\max_{\bm{\Theta},\bm{\Pi}, \mathbf{v}} \sum_{i=1}^{N} v_{i} \left \{ \log p_{\mathcal{F}}\left(y_i|\mathbf{x}_i,\bm{\Theta},\bm{\Pi} \right) + \gamma H_i\right \}  + \zeta \sum_{i=1}^N \log\left(v_i + \zeta/\lambda\right) ,
\end{equation}
where $\zeta$ is a parameter controlling the learning pace.
We set $\zeta=\left(\frac{1}{\lambda'}-\frac{1}{\lambda}\right)^{-1}$, and $\lambda>\lambda'>0$ to construct a reasonable soft weighting formulation.
The self-paced regularizer in Eq.~(\ref{Eq.9}) is convex with respect to $v\in\left[0,1\right]$.
Then, setting the partial gradient of Eq.~(\ref{Eq.9}) with respect to $v_i$ as zero will lead to the following:
\begin{equation}
\log p_{\mathcal{F}}\left(y_i|\mathbf{x}_i,\bm{\Theta},\bm{\Pi} \right) + \gamma H_i + \frac{\zeta}{v_i + \zeta/\lambda} = 0.
\end{equation}
Then, the optimal solution of $v_i$ is given by:
\begin{align}
v_i^* = \left\{ \begin{array}{ll}
1 & \textrm{if $\log p_{\mathcal{F}i} + \gamma H_i \geq -\lambda'  $}\\
0 & \textrm{if $\log p_{\mathcal{F}i} + \gamma H_i \leq -\lambda $}\\
\frac{-\zeta}{\log p_{\mathcal{F}i} + \gamma H_i} - \zeta/\lambda & \textrm{otherwise}
\end{array} \right.
\label{Eq.11}
\end{align}
If either the log likelihood or the predictive uncertainty is too large,  $v^*_i$ equals to 1.
In addition, if the likelihood and the predictive uncertainty are both too small, $v^*_i$ equals to 0.
Except the above two situations, the soft weighting calculation (\ie, the last line of Eq.~(\ref{Eq.11})) is adopted.

\noindent \textbf{Curriculum Reconstruction.}
The underrepresented examples play an important role in our SPUDRFs algorithm.
As previously mentioned, the proposed new self-paced regime coupled with a mixture weighting scheme emphasizes more on underrepresented examples, rendering better solutions.
Since the intrinsic reason that causes predictive uncertainty is plausibly the imbalanced training data, we further re-balance data distribution via a curriculum reconstruction strategy.
More specifically, we distinguish the underrepresented examples (whose $H_i$ is lager than $\beta$) from regular ones at each pace, and augment them into the training data.

\subsection{Optimization}
\label{Learning}
We propose a two-step alternative search strategy (ASS) algorithm to solve SPUDRFs: (\romannumeral1) update $\mathbf{v}$ for sample selection with fixed $\bm{\Theta}$ and $\bm{\Pi}$, and (\romannumeral2) update $\bm{\Theta}$ and $\bm{\Pi}$ with current fixed sample weights $\mathbf{v}$.

\noindent\textbf{Optimizing $\bm{\Theta}$ and $\bm{\Pi}$.}
The parameters $\left\{\bm{\Theta},\bm{\Pi}\right\}$ and weights $\mathbf{v}$ are optimized alternatively.
With fixed $\mathbf{v}$,  our DRFs is learned by alternatively updating $\bm{\Theta}$ and $\bm{\Pi}$.
In \cite{shen_deep_2018}, the parameters $\bm{\Theta}$ for split nodes (\ie~parameters for VGG) are updated through gradient descent since the loss is differentiable with respect to $\bm{\Theta}$.
While the parameters $\bm{\Pi}$ for leaf nodes are updated by virtue of variational bounding~\cite{shen_deep_2018} when fixing $\bm{\Theta}$.

\noindent\textbf{Optimizing $\mathbf{v}$.}
As previously discussed, $v_i$ is a binary variable or real variable ranged in $\left[0, 1\right]$.
It indicates how to weight the $i^{th}$ sample during training.
The parameter $\lambda$ could be initialized to obtain 50\% samples to train the model, and is then progressively increased to involve 10\% more data in each pace.
The parameter $\gamma$ could be initialized empirically and is progressively decayed to zero.
The training stops when all the samples are selected, at $\gamma=0$.
Along with increasing $\lambda$ and decreasing $\gamma$, DRFs are trained to be more ``mature''.
This learning process is like how our human beings learn one thing from easy and uncertain to complex.

\section{Experimental Results}
\subsection{Tasks and Benchmark Datasets}
\noindent \textbf{Age Estimation.}
The Morph \uppercase\expandafter{\romannumeral2~\cite{ricanek2006morph}} dataset contains 55,134 unique face images of 13618 individuals with unbalanced gender and ethnicity distributions, and is the most popular publicly available real age dataset.
The FG-NET~\cite{panis2016overview} dataset includes 1,002 color or gray images of 82 people with each subject almost accompanied by more than 10 photos at different ages.
Since all images were taken in a totally uncontrolled environment, there exists a large deviation on lighting, pose and expression (\ie~PIE) of faces inside the dataset.

\noindent \textbf{Head Pose Estimation.}
The BIWI dataset~\cite{fanelli2013random} contains 20 subjects, of which 10 are male and 6 are female, besides, 4 males have been chosen twice with wearing glasses or not.
It includes 15678 images collected by a Kinect sensor device for different persons and head poses
with pitch, yaw and roll angles mainly ranging within $\pm 60^{\circ}$, $\pm 75^{\circ}$ and $\pm 50^{\circ}$.

\subsection{Experimental Setup}

\noindent\textbf{Dataset Setting.}
The settings of different datasets are given below.
\begin{itemize}
	\item \textbf{Morph \uppercase\expandafter{\romannumeral2}.} Following the recent relevant work~\cite{shen_deep_2018}, the images in Morph \uppercase\expandafter{\romannumeral2} were divided into two sets: 80\% for training and the rest 20\% for testing. The random division was repeated
	5 times and the reported performance was averaged over these 5 times. The VGG-Face~\cite{parkhi2015deep} networks were chosen as the pre-trained model.
	\item \textbf{FG-NET.} The leave-one-person-out scheme~\cite{shen_deep_2018} was adopted, where the images of one person were selected for testing and the remains for training. The VGG-16 networks were pre-trained on the IMDB-WIKI~\cite{rothe2018deep} dataset.
	\item \textbf{BIWI.} Similarly, 80\% of the whole data was randomly chosen for training and the rest 20\% for testing, and this operation was repeated 5 times. Moreover, the VGG-FACE networks were the pre-trained model.
\end{itemize}

\noindent\textbf{Evaluation Metrics.}
The first evaluation metric is the mean absolute error (MAE), which is defined as the average absolute error between the ground truth and the predicted output: $\sum_{i=1}^{N}\left|\hat{y}_{i}-y_{i}\right|/N$, $\hat{y_{i}}$ represents the estimated output of the $i^{th}$ sample, and $N$ is the total number of testing images.
The other evaluation metric is cumulative score (CS), which denotes the percentage of images sorted in the range of $\left[y_{i}-L, y_{i}+L\right]$: $CS(L)=\sum_{i=1}^{N}\left[\vert\hat{y}_{i}-y_{i}\vert \leq L\right]/N \cdot 100 \%$, where $[ \cdot ]$ denotes an indicator function and $L$ is the error range.

\noindent\textbf{Preprocessing and Data Augmentation.}
On the Morph \uppercase\expandafter{\romannumeral2} and FG-NET datasets, MTCNN~\cite{zhang_joint_2016} was used for joint face detection and alignment.
Furthermore, following~\cite{shen_deep_2018}, we augmented training images in three ways: (\romannumeral1) random cropping (5 times); (\romannumeral2) adding Gaussian white noise with variance of 0.0001 (2 times); (\romannumeral3) random horizontal flipping (2 times). The whole number of samples was increased by 20 times after augmentation.
On the BIWI dataset, we utilized the depth images for training and did not augment training images.

\noindent\textbf{Parameters Setting.}
The VGG-16~\cite{Simonyan2015} was employed as the fundamental backbone networks of SPUDRFs.
The hyper-parameters of VGG-16 were: training batch size (32 on Morph \uppercase\expandafter{\romannumeral2} and BIWI, 8 on FG-NET), drop out ratio (0.5), max iterations of each pace ($80k$ on Morph \uppercase\expandafter{\romannumeral2},  $20k$ on FG-NET, and $40k$ on BIWI), stochastic gradient descent (SGD), initial learning rate (0.2 on Morph \uppercase\expandafter{\romannumeral2}, 0.1 on BIWI, 0.02 on FG-NET) by reducing the learning rate ($\times$0.5) per $10k$ iterations. The hyper-parameters of SPUDRFs were: tree number (5), tree depth (6), output unit number of feature learning (128), iterations to update leaf node predictions (20), number of mini-batches used to update leaf node predictions (50).
In the first pace, 50\% samples which are easy or underrepresented were selected for training.
Here, $\lambda$ was set to guarantee the first 50\% samples with large $\log p_{\mathcal{F}i} + \gamma H_i$ values  involved.
$\lambda'$ was set to ensure 10\% of selected samples with soft weighting.
$\gamma$ was initialized to be 15 on the Morph \uppercase\expandafter{\romannumeral2} and BIWI datasets, and 5 on the FG-NET dataset.
$\beta$ was set to select 1180 and 2000 samples as the ones needed to be augmented twice at each pace on the Morph \uppercase\expandafter{\romannumeral2} and BIWI datasets.
The number of paces was empirically set to be 10, 3 and 6 on the Morph \uppercase\expandafter{\romannumeral2}, FG-NET, and BIWI datasets, and except the first pace, an equal proportion of the rest data was gradually involved at each pace.

\subsection{Validity of Our Proposed Method}
\label{valid}
\begin{figure}[t]
	\centering
	\includegraphics[width=0.99\textwidth]{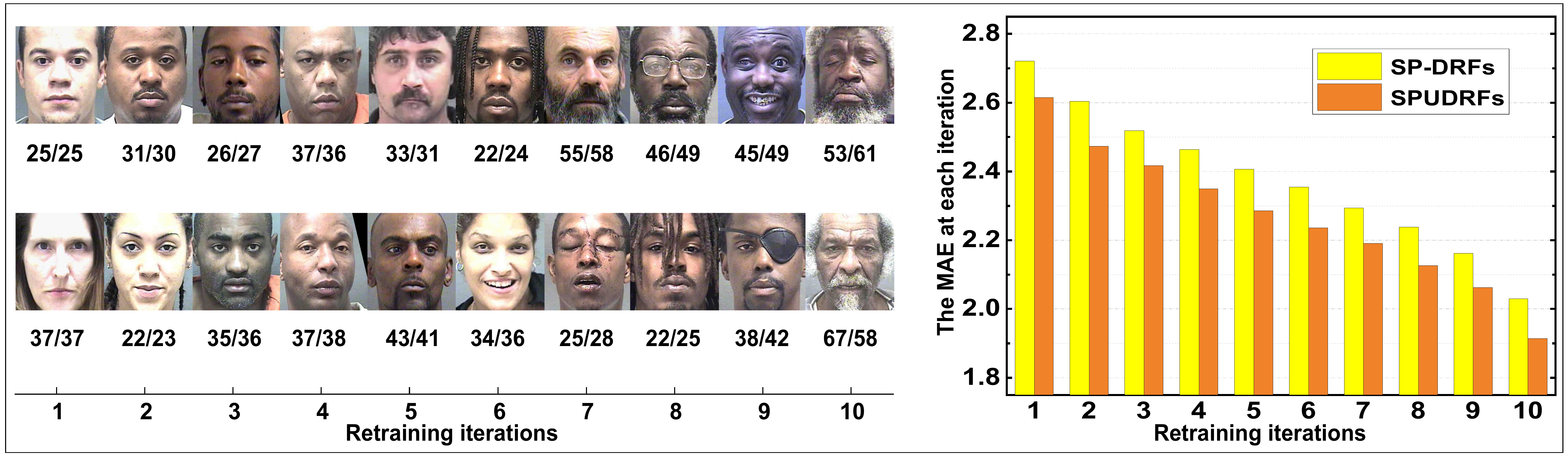}
	\caption{The gradual learning process of SP-DRFs and SPUDRFs. \textbf{Left:} The typical worst cases at each iteration become more confusing and noisy along
		with iteratively increasing $\lambda$ and decreasing $\gamma$. The two numbers below each image are the real age (left) and predicted age (right). \textbf{Right:} The MAEs of SP-DRFs and SPUDRFs at each pace descend gradually. The SPUDRFs show its superiority of taking predictive uncertainty into consideration, when compared with SP-DRFs.}
	\label{SPUDRFs_validation}
\end{figure}

\noindent \textbf{Self-paced Learning Strategy.}
The validity of self-paced strategy in training DDMs is mainly demonstrated by the following experiments on the MorphII dataset.
We first used all training images in the Morph \uppercase\expandafter{\romannumeral2} datasets to train DRFs so as to rank samples at the beginning pace.
Retraining proceeded with progressively increasing $\lambda$ such that every 1/9 of the rest data was gradually involved at each pace, where $\gamma$ was decreased to the half of its previous value every time.
In the last pace, the value $\gamma$ was constrainedly set to be 0.
The visualization of this process can be found in Fig.~\ref{SPUDRFs_validation}.

Fig.~\ref{SPUDRFs_validation} illustrates the representative face images in each learning pace of SPUDRFs, along with increasing $\lambda$ and decreasing $\gamma$.
The two numbers below each image are the real age (left) and predicted age (right).
We observe that the training images in the latter paces are obviously more confusing and noisy than the ones in the early paces.
Since our model is initialized by the results of the previous retraining pace, meaning adaptively calibrated by ``good'' examples.
As a result, it has improved performance than DRFs, where the MAE is improved from 2.17 to 1.91, and the CS is promoted from 92.79\% to 93.31\% (see Fig.~\ref{morph_experiment}(a)).

Fig.~\ref{SPUDRFs_validation} also shows the comparison between SP-DRFs and SPUDRFs on the Morph \uppercase\expandafter{\romannumeral2} datasets.
The yellow bar denotes the MAE of SP-DRFs, while the orange bar denotes for SPUDRFs.
We find the MAE of SPUDRFs is lower than SP-DRFs at each pace, particularly the last pace ($1.91$ against $2.02$).
As we discussed previously, as in Fig.~\ref{Figure1}, SPUDRFs are prone to reach less biased solutions due to the wider covering range of leaf nodes, owing to considering underrepresented examples.
This experiment could be regarded as an ablation study of considering ranking fairness in SPL.

\begin{figure}
	\centering
	\includegraphics[width=0.85\textwidth]{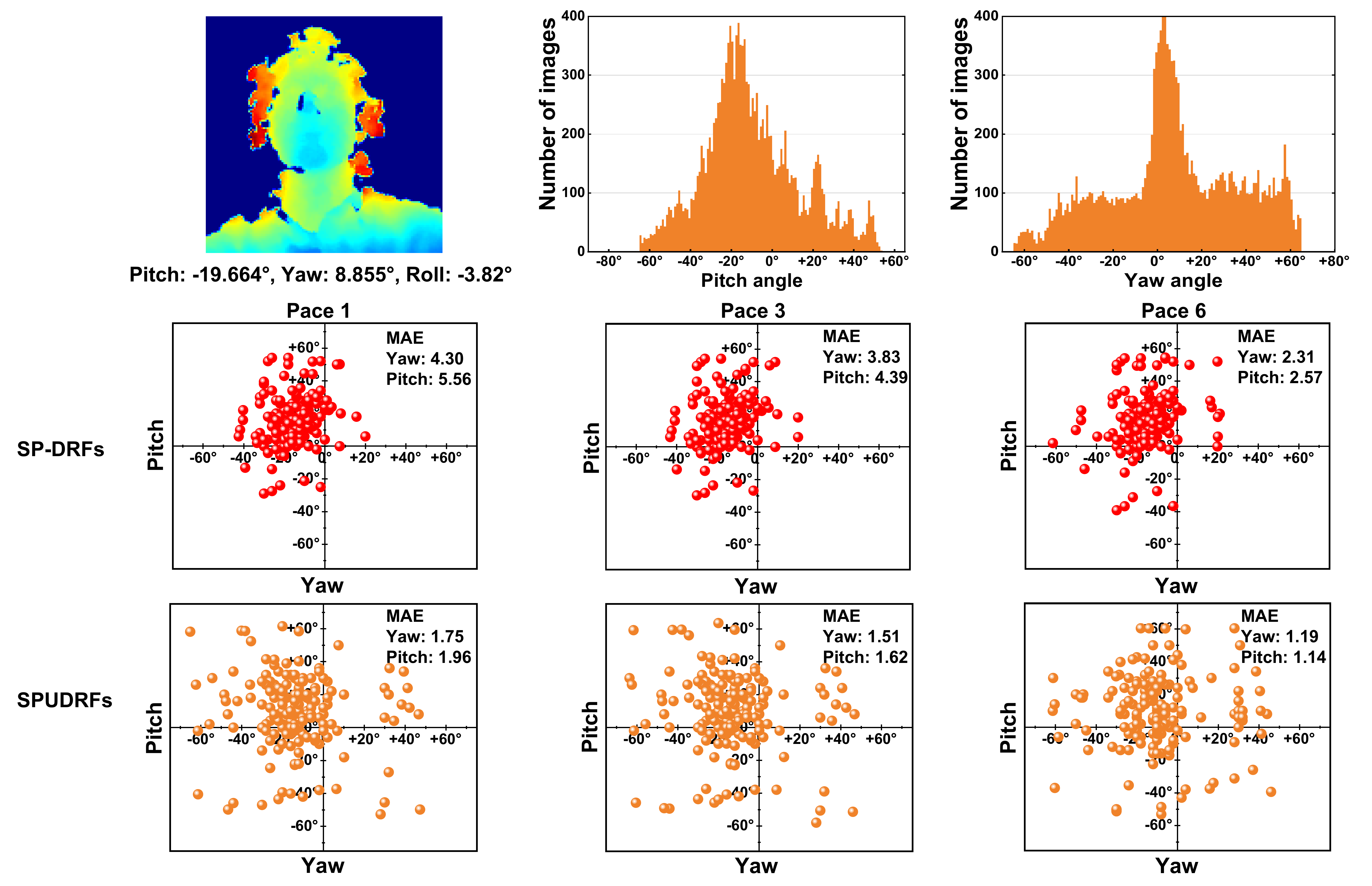}
	\caption{The leaf node distribution of SP-DRFs and SPUDRFs in gradual learning process. Three paces, \ie~pace 1, 3, and 6, are randomly chosen for visualization. For SP-DRFs, the Gaussian means of leaf nodes (the red points in the second row) are concentrated in a small range, incurring seriously biased solutions. For SPUDRFs, the Gaussian means of leaf nodes (the orange points in the third row) distribute widely, leading to much better MAE performance.}
	\label{Uncertainty_efficacy}
\end{figure}
\noindent\textbf{Considering Underrepresented Examples.}
\label{ExpUnderSamples}
On the BIWI dataset, the necessity of considering ranking fairness in SPUDRFs is further demonstrated.
In SP-DRFs, DRFs was first trained on the basis of all data, and the samples were ranked and selected for the first pace according to this result.
Subsequently, every $10\%$ of the rest samples were progressively involved for retraining.
$\lambda$ was progressively increased while $\gamma$ was progressively decreased until zero.
In SP-DRFs, the same self-paced strategy was adopted as in SPUDRFs, but without considering ranking fairness (\ie~underrepresented examples).

Fig.~\ref{Uncertainty_efficacy} visualizes the leaf node distributions of SP-DRFs and SPUDRFs in the progressive learning process.
The Gaussian means $\mu_l$ associated with the 160 leaf nodes, where each 32 leaf nodes are defined for 5 trees, are plotted in each sub-figures.
Three paces, \ie~pace 1, 3, and 6, are randomly chosen for visualization.
Only pitch and yaw angles are shown for clarity.
Besides, the distribution of angle labels (\ie~pitch and yaw) are also shown, where the imbalance problem of data distribution is obvious.

In Fig.~\ref{Uncertainty_efficacy}, the comparison results between SP-DRFs and SPUDRFs demonstrate the efficacy of considering ranking fairness in SPL.
For SP-DRFs, the Gaussian means of leaf nodes (red points in the second row) are concentrated in a small range, incurring seriously biased solutions.
That means the underrepresented examples have been neglected in SPL training.
The poor MAEs are the evidence for this, which are even inferior to DRFs (see Fig.~\ref{biwi_experiment}(a)).
SPUDRFs rank samples by log likelihood coupled with entropy, and are prone to achieve less biased solutions,  as shown in the third rows of Fig.~\ref{Uncertainty_efficacy}.
Such an experiment could be also regard as an ablation study of the proposed ranking algorithm.

\subsection{Comparison with State-of-the-art Methods}
\label{sec:blind}

We compared our SPUDRFs with other state-of-the-art methods on the Morph \uppercase\expandafter{\romannumeral2}, FG-NET and BIWI datasets.
\begin{figure}[t] 
	\centering                                                                         
	\begin{tabular}[h]{cc}
		\small
		\scalebox{0.82}{
			\begin{tabular}{@{}l|c|c}
				\hline
				Method & MAE$\downarrow$ & CS$\uparrow$\\
				\hline
				\hline
				LSVR \cite{guo_human_2009}     & 4.31 & 66.2\% \\
				RCCA \cite{Huerta2014Facial}   & 4.25 & 71.2\% \\
				OHRank \cite{Chang2011Ordinal} & 3.82 & N/A \\
				OR-CNN \cite{niu_ordinal_2016} & 3.27 & 73.0\% \\
				Ranking-CNN \cite{chen_using_2017} & 2.96 & 85.0\% \\
				DRFs \cite{shen_deep_2018} & 2.17 & 91.3\% \\
				DLDL-v2 \cite{gao_age_2018}& 1.97 & N/A \\
				\textbf{SP-DRFs} & \textbf{2.02} & \textbf{92.79\%} \\
				\textbf{SPUDRFs} & \textbf{1.91} & \textbf{93.31\%} \\
				\hline
			\end{tabular}
		}
		& \raisebox{-0.83in}{\includegraphics[width=0.555\textwidth]{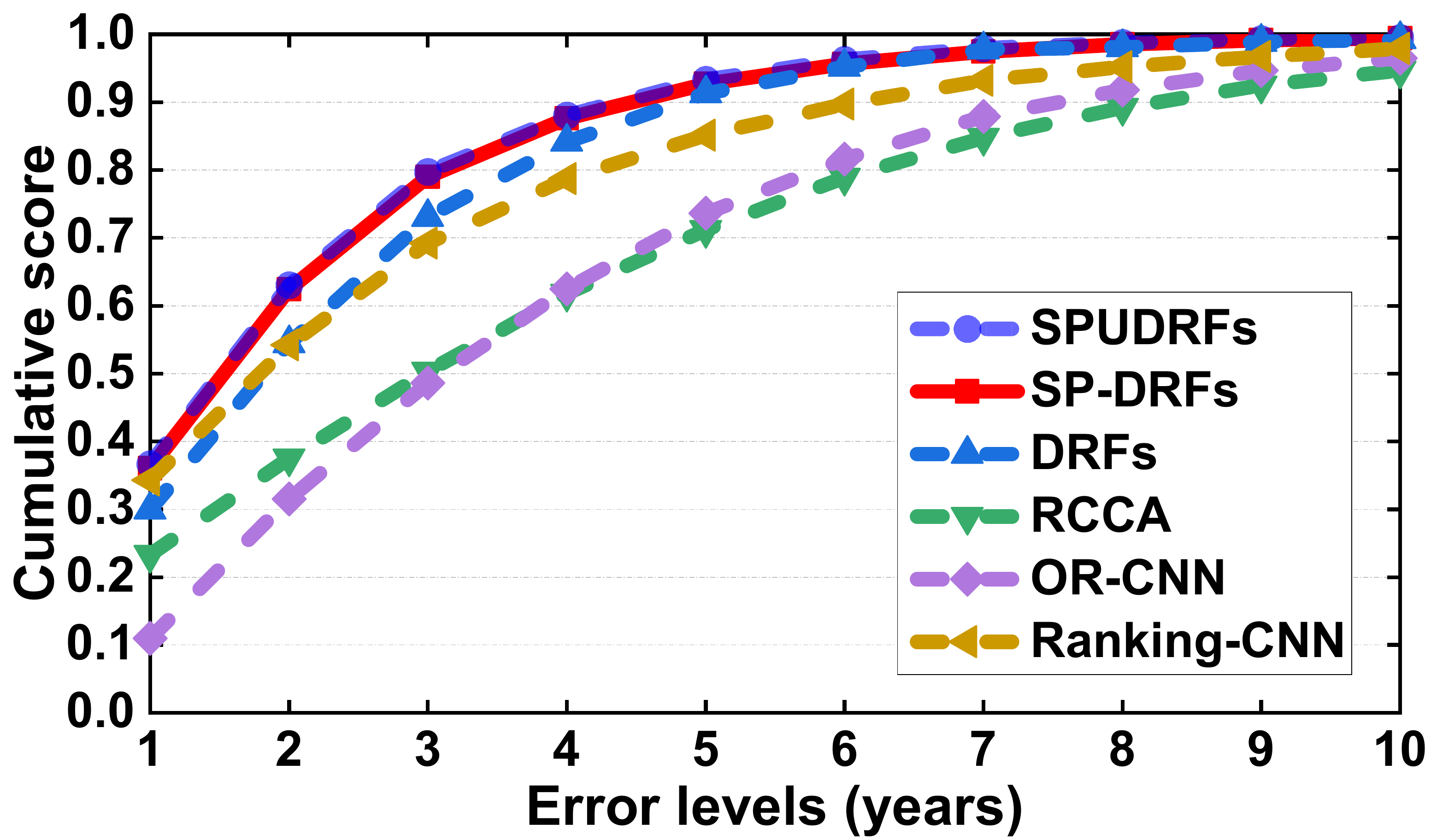}} \\
		{\small (a) } & {\small (b)}
	\end{tabular}
	\caption{The comparison results on the Morph \uppercase\expandafter{\romannumeral2} dataset. (a) The MAE comparison with the state-of-the-art methods, (b) the CS curves of the comparison methods.}
	\label{morph_experiment}
\end{figure}

\noindent\textbf{Results on Morph \uppercase\expandafter{\romannumeral2}.}
Fig. \ref{morph_experiment}(a) compares SPUDRFs with other baseline methods: LSVR~\cite{guo_human_2009}, RCCA \cite{Huerta2014Facial}, OHRank~\cite{Chang2011Ordinal}, OR-CNN \cite{niu_ordinal_2016}, Ranking-CNN \cite{chen_using_2017}, DRFs~\cite{shen_deep_2018}, and DLDL-v2~\cite{gao_age_2018}.
Firstly, owing to the effective feature learning ability of DNNs, the SPUDRFs method is much superior to the shallow model based approaches, such as LSVR~\cite{guo_human_2009} and OHRank~\cite{Chang2011Ordinal}.
Secondly, duing to the valid self-paced regime, our SPUDRFs outperform other DDMs, and lead to more robust and less biased solutions.
Thirdly, SPUDRFs outperform SP-DRFs on both MAE and CS, and achieve state-of-the-art performance.
Fig.~\ref{morph_experiment}(b) shows the CS comparison on this dataset.
We observe that the CS of SPUDRFs reachs 93.31\% at error level $L=5$, which is significantly better than DRFs and obtained 2.01\% increment.
\begin{figure} 
	\centering                                                                         
	\begin{tabular}[h]{cc}
		\small
		\scalebox{0.82}{
			\begin{tabular}{@{}l|c|c}
				\hline
				Method & MAE$\downarrow$ & CS$\uparrow$\\
				\hline
				\hline
				IIS-LDL \cite{xin_geng_facial_2013} & 5.77 & N/A \\
				LARR \cite{guodong_guo_image-based_2008} & 5.07 & 68.9\% \\
				MTWGP \cite{Yu2010Multi} & 4.83 & 72.3\% \\
				DIF \cite{han_demographic_2015} & 4.80 & 74.3\% \\
				OHRank \cite{Chang2011Ordinal} & 4.48 & 74.4\% \\
				CAM \cite{Luu2013Contourlet} & 4.12 & 73.5\% \\
				DRFs \cite{shen_deep_2018} & 3.06 & 83.33\% \\
				\textbf{SP-DRFs} & \textbf{2.84} & \textbf{84.73\%} \\
				\textbf{SPUDRFs} & \textbf{2.77} & \textbf{85.53\%}\\
				\hline
			\end{tabular}
		}
		
		& \raisebox{-0.83in}{\includegraphics[width=0.555\textwidth]{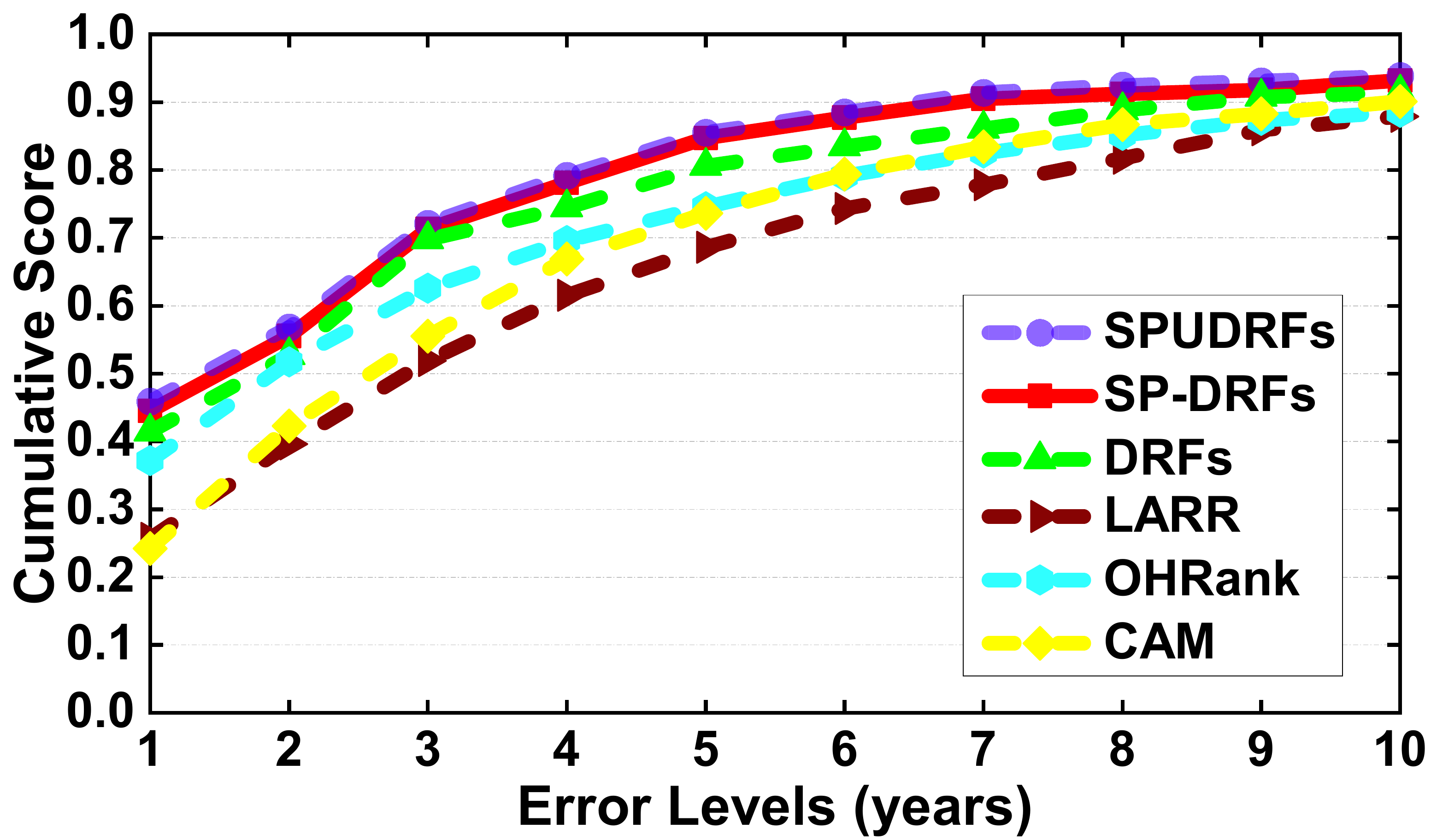}} \\
		{\small (a) } & {\small (b)}
	\end{tabular}
	\caption{The comparison results on the FGNET dataset. (a) The MAE comparison with the state-of-the-art methods, (b) the CS curves of the comparison methods.}
	\label{fgnet_experiment}
\end{figure}

\noindent\textbf{Results on FG-NET.} Fig.~\ref{fgnet_experiment}(a) shows the comparison results of SPUDRFs with the state-of-the-art approaches on FG-NET dataset.
As can be seen, SPUDRFs reach an MAE of 2.77 years, which reduces the MAE of DRFs by 0.29 years.
Besides, the CS comparison is shown in Fig.~\ref{fgnet_experiment}(b), SPUDRFs consistently outperform other recent proposed methods at different error levels, proving that our method is effective in enhancing the robustness of facial age estimation.

\begin{figure} 
	\centering                                                                         
	\begin{tabular}[h]{cc}
		\small
		\scalebox{0.78}{
			\begin{tabular}{@{}l|c}
				\hline
				Method & MAE$\downarrow$\\
				\hline
				\hline
				HF \cite{riegler2013hough} & 4.95 \\
				SVR \cite{drucker1997support} & 3.14 \\
				RRF \cite{liaw2002classification} & 3.06 \\
				KPLS \cite{al2012partial} & 2.88 \\
				SAE \cite{hinton2006reducing} & 1.94 \\
				MoDRN \cite{huang2018mixture} & 1.62 \\
				DRFs \cite{shen_deep_2018} & 1.44 \\
				\textbf{SP-DRFs} & \textbf{2.08} \\
				\textbf{SPUDRFs} & \textbf{1.18} \\
				\hline
			\end{tabular}
		}
		& \raisebox{-0.8in}{\includegraphics[width=0.53\textwidth]{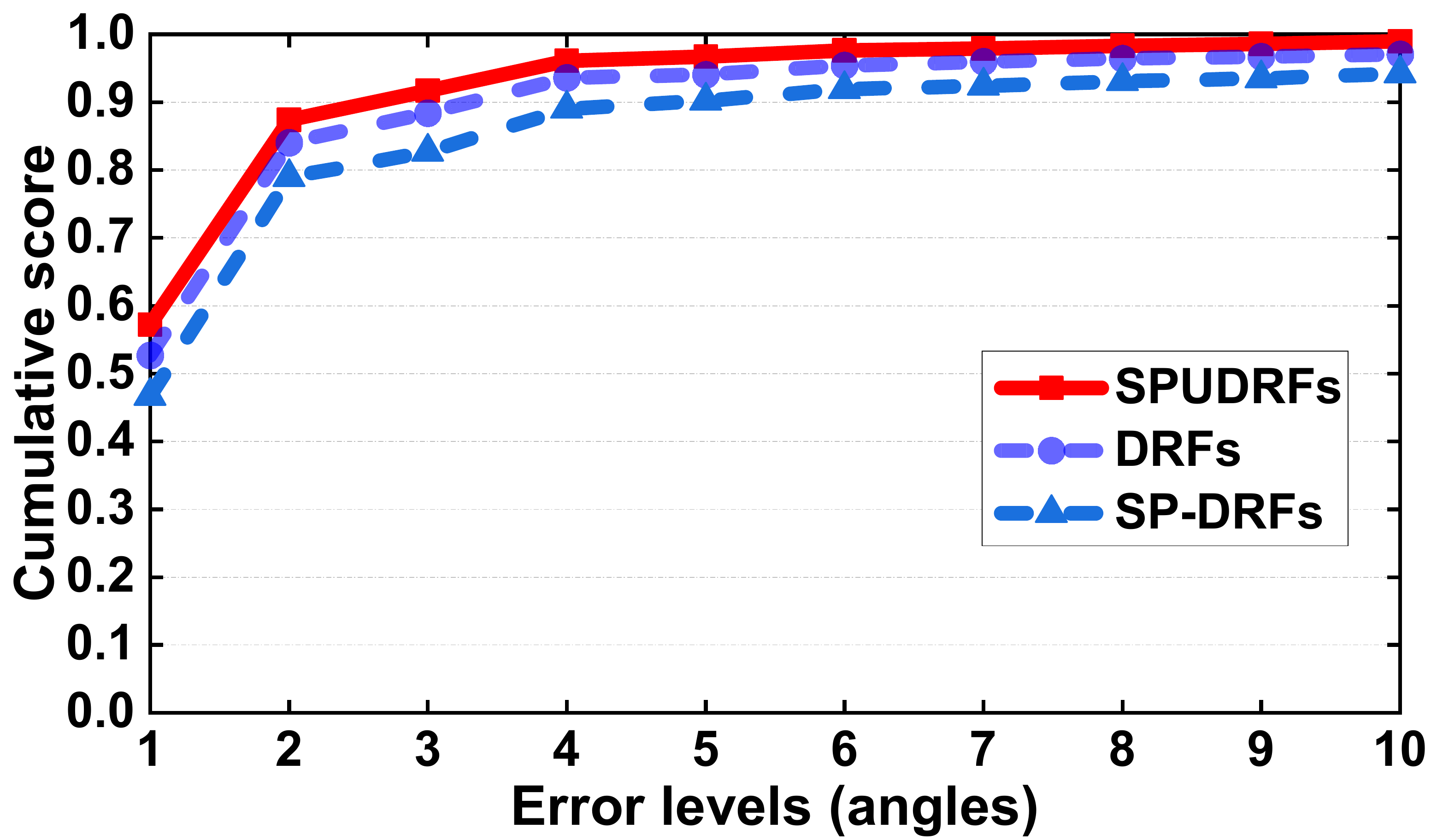}} \\
		{\small (a) } & {\small (b)}
	\end{tabular}
	\caption{The comparison results on the BIWI dataset. (a) The MAE comparison with the state-of-the-art methods, (b) the CS curves the comparison methods.}
	\label{biwi_experiment}
\end{figure}

\noindent\textbf{Results on BIWI.}
Fig.~\ref{biwi_experiment}(a) shows the comparison results of our method with several state-of-the-art approaches.
The experimental results reveal the proposed SPUDRFs method achieves the best performance with an MAE of 1.18, which is state-of-the-art.
\emph{Besides, we observe one important phenomenon: the MAE of SP-DRFs is even much worse than DRFs.
	This further demonstrates the obvious drawback of the ranking and selecting algorithm in original SPL---incurring seriously biased solutions.}
In the first pace of the original SPL, as illustrated in Fig.~\ref{Uncertainty_efficacy}, the Gaussian means of leaf nodes are concentrated in a small range, leading biased solutions.
Incorporating underrepresented examples in the early pace of SPUDRFs renders to more reasonable distributions of the leaf nodes.
Fig.~\ref{biwi_experiment}(b) plots only three CS curves for brevity, \ie, DRFs, SP-DRFs and SPUDRFs, which is the average of the three angles.
SPUDRFs also outperform DRFs and SP-DRFs at different error levels.

\section{Conclusion and Future Work}
This paper explored how self-paced regime leads deep discriminative models (DDMs) to achieve more robust and less biased solutions on different computer vision tasks (\eg~facial age estimation and head pose estimation).
Specifically, a novel self-paced paradigm, which considers ranking fairness, was proposed.
The new ranking scheme jointly considers loss and predictive uncertainty.
Such a paradigm was combined with deep regression forests (DRFs), and led to a new model, namely self-paced deep regression forests with consideration on underrepresented examples (SPUDRFs).
Extensive experiments on two well-known computer vision tasks demonstrated the efficacy of the proposed paradigm.

We are currently applying self-paced DDMs for other computer vision tasks, \eg~viewpoint estimation, indoor scene classification, where the ability to handle ranking unfairness is fundamental to the success.
Thus, investigating the causes of algorithm unfairness in DDMs is a worthy direction.
Obviously, except data imbalance, there exist some other causing factors.
In addition to this, exploring how to combine the new self-paced paradigm with other DDMs, including deep regressors and  classifiers, will also be our future work.

\noindent \textbf{Acknowledgement.} The authors gratefully acknowledge the support of China Postdoctoral Science Foundation No.2017M623007.

\clearpage
%
%
\bibliographystyle{splncs04}
\bibliography{reference}
\end{document}